\documentclass[letterpaper]{article} % DO NOT CHANGE THIS
\usepackage{aaai2026}  % DO NOT CHANGE THIS
\usepackage{times}  % DO NOT CHANGE THIS
\usepackage{helvet}  % DO NOT CHANGE THIS
\usepackage{courier}  % DO NOT CHANGE THIS
\usepackage[hyphens]{url}  % DO NOT CHANGE THIS
\usepackage{graphicx} % DO NOT CHANGE THIS
\usepackage{natbib}  % DO NOT CHANGE THIS AND DO NOT ADD ANY OPTIONS TO IT
\usepackage{caption} % DO NOT CHANGE THIS AND DO NOT ADD ANY OPTIONS TO IT
\usepackage{algorithmic}
\usepackage{booktabs}
\usepackage{multirow}

\usepackage{listings}
\usepackage{subcaption}
\DeclareCaptionStyle{ruled}{labelfont=normalfont,labelsep=colon,strut=off} % DO NOT CHANGE THIS
\usepackage{amsmath, amssymb}
\usepackage[linesnumbered,ruled,vlined]{algorithm2e}
\title{CoRe-Fed: Bridging Collaborative and Representation Fairness via \\Federated Embedding Distillation}

\author{
    Noorain Mukhtiar\textsuperscript{\rm 1},  Adnan Mahmood\textsuperscript{\rm 1}, Quan Z. Sheng\textsuperscript{\rm 1}\\
}
\affiliations{
    \textsuperscript{\rm 1}School of Computing, Macquarie University, Sydney, NSW 2109, Australia\\
    noorain.mukhtiar@hdr.mq.edu.au, {\{adnan.mahmood, michael.sheng\}@mq.edu.au}
}

\begin{document}

\maketitle

\begin{abstract}

With the proliferation of distributed data sources, Federated Learning (FL) has emerged as a key approach to enable collaborative intelligence through decentralized model training while preserving data privacy. However, conventional FL algorithms often suffer from performance disparities across clients caused by heterogeneous data distributions and unequal participation, which leads to unfair outcomes. Specifically, we focus on two core fairness challenges, i.e., \emph{representation bias}, arising from misaligned client representations, and \emph{collaborative bias}, stemming from inequitable contribution during aggregation, both of which degrade model performance and generalizability. To mitigate these disparities, we propose CoRe-Fed, a unified optimization framework that bridges collaborative and representation fairness via embedding-level regularization and fairness-aware aggregation. 
Initially, an alignment-driven mechanism promotes semantic consistency between local and global embeddings to reduce representational divergence. Subsequently, a dynamic reward-penalty-based aggregation strategy adjusts each client’s weight based on participation history and embedding alignment to ensure contribution-aware aggregation. Extensive experiments across diverse models and datasets demonstrate that CoRe-Fed improves both fairness and model performance over the state-of-the-art baseline algorithms.

\end{abstract}

\begin{links}
\link{Code}{https://github.com/Noorain1/CoRe-Fed}
\end{links}

\section{Introduction}
\label{Sec: Introduction}
Federated Learning (FL) has gained widespread adoption as a decentralized learning framework that enables multiple devices, sensors, or edge nodes (collectively referred to as clients or participants) to collaboratively train Machine Learning (ML) models without directly sharing their raw data, thus preserving data privacy \cite{ijcai2024p919}. However, despite its growing popularity, traditional FL encounters several challenges, particularly, in ensuring fair and unbiased model performance across clients with heterogeneous data distribution and varying participation frequency. These discrepancies result in models that perform disproportionately well on certain clients, while neglecting others, a phenomenon referred to as performance bias. 

Such bias in FL can transpire at various stages of the training process, significantly undermining the reliability and overall performance of the resulting global model \cite{AAAI_2021_model_performance}. An unfair model can lead to several adverse outcomes, i.e., misrepresentation of client-specific data distributions, over- or under-fitting, marginalizing certain clients, and making suboptimal decisions. To this end, we target 
the following 
three key biasing challenges in FL.

\begin{figure}[t]
    \centering
    \includegraphics[width=1\linewidth]{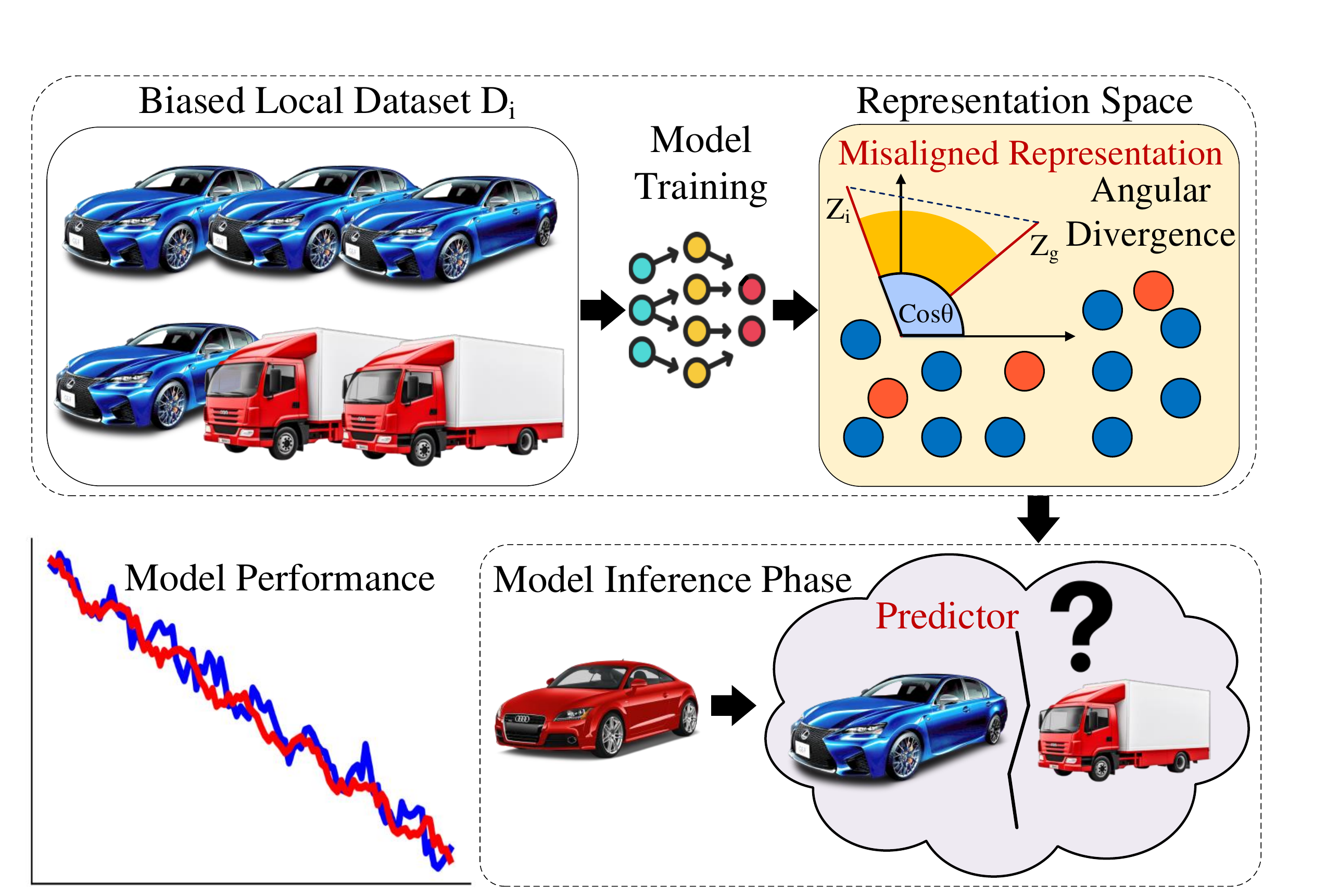}
    \caption{Illustration of representation bias in FL using CIFAR-10 automobiles images. Clients trained on biased local data, e.g., cars vs. trucks, leads to misaligned feature representations in the global embedding space. As shown in the upper section, angular divergence emerges when client representations deviate from semantically correct directions. This misalignment causes degraded performance, particularly, during inference on bias-conflicting samples.
    }
    \label{fig:intro}
\end{figure}
    \label{Fig:1}
\textbf{Performance Bias --} 
In practical scenarios, client datasets often exhibit statistical heterogeneity \cite{ye2023heterogeneous} and label correlations \cite{Label_COrr_2025_trans} that introduce bias into the learning process. In particular, biased datasets \cite{ DNN_Bias_NIPS_2020} contain features that strongly correlate with class labels in the training dataset but do not accurately capture the inherent semantic representations. It is pertinent to mention that Deep Neural Network (DNN) models trained with such biased data are more prone to base decisions on bias-related attributes rather than meaningful representations. This, in turn, results in substantial performance bias and accuracy drop during inference, particularly when a model encounters bias-conflicting samples \cite{Model_performance_bias}.
 
\textbf{Representation Bias --} 
In federated settings, clients with underrepresented classes or non-IID data tend to produce embeddings that poorly align with the global model’s latent space. When a global model is trained
on these locally biased datasets, it tends to form biased representations \cite{Zhang}. This phenomenon entails the emergence of client-specific clusters within a model’s learned representations, reflecting the statistical disparities across clients. As illustrated in Figure \ref{Fig:1}, client embeddings are scattered and poorly aligned, even for samples belonging to the same class, indicating that the global model struggles to learn a unified, semantically meaningful representation.

\textbf{Collaborative Bias --}
While addressing representation disparity is critical for learning meaningful and generalizable features across clients, we argue that it is not sufficient to guarantee fairness in federated settings. 
Representation alignment improves feature space consistency across clients, however, bias can still emerge during aggregation if client representations are aggregated regardless of their contribution to a global model. Standard aggregation strategies, i.e., FedAvg \cite{mcmahan2017communication}, may (a) inadvertently suppress meaningful updates from underrepresented clients, (b) neglect infrequently participating clients, or (c) dilute high-quality representations contributed by well-performing clients, thereby compromising overall representation fairness. In contrast, assigning equal weight to all updates allows noisy or misaligned contributions to influence a global model to the same extent as well-aligned contributions \cite{IJCAI_Shapley, Colab_Fair_KDD}, thereby leading to collaborative bias, i.e., a systemic bias rooted in the aggregation process. Such bias not only compromises fairness but also discourages participation from marginalized clients, potentially triggering a feedback loop that further weakens their influence. This, in turn, undermines the global model’s ability to generalize across diverse data distributions and leads to performance degradation.

To address the aforementioned challenges, we propose CoRe-Fed (\textbf{Co}llaborative \textbf{Re}presentation Fairness in \textbf{Fed}erated Learning), a novel framework that jointly mitigates representation and collaborative biases to reduce model performance disparities. CoRe-Fed introduces two key mechanisms, i.e., embedding-level alignment-aware optimization and contribution-aware aggregation. The former employs a contrastive embedding alignment strategy based on knowledge distillation to align local client representations with a global embedding prototype. This ensures clients' semantic consistency in a shared latent space regardless of their local data skew. The latter applies a contribution-aware aggregation scheme to adjust client weights based on a combination of negative participation frequency and embedding representation similarity. 

It is worth mentioning that these two fairness notions are mutually reinforcing, i.e., representation fairness enhances the quality of client embeddings, whereas collaborative fairness ensures that those improved embeddings are not diluted during aggregation process. To the best of our knowledge, CoRe-Fed is among the first to explicitly bridge representation and collaborative fairness in a unified framework. Through comprehensive experimentation with the state-of-the-art methods, we demonstrate that mitigating bias in both the feature space and the aggregation process leads to more equitable and generalizable models in heterogeneous federated environments. Our main contributions are as follows:

\begin{itemize}

\item We propose CoRe-Fed, a novel FL framework that jointly maintains representation and collaborative fairness through embedding-level alignment and contribution-aware aggregation.

\item We introduce a contrastive learning-based embedding alignment strategy that leverages knowledge distillation to align local client representations with a global embedding prototype, thereby promoting semantic consistency across heterogeneous clients.

\item We design a contribution-aware aggregation mechanism that adjusts client weights based on their participation frequency and representational alignment, thereby amplifying the influence of semantically rich and underrepresented clients.

\item To validate the effectiveness of CoRe-Fed, we conduct extensive experiments on different FL scenarios. Our experimental results indicate substantial improvements in fairness and comparable accuracy with the state-of-the-art methods. 

\end{itemize}

\section{Background and Related Work} 
\label{Sec: Background and Related Work}
\subsection{Layer-Wise Fair Federated Learning}
As fairness has gained significant momentum in recent years, several studies have adopted layer-wise approaches to mitigate bias in FL.  For instance, \cite{ICLR2025_LayerwiseFL} proposes an adaptive layer-wise weight shrinking step after model aggregation. \cite{Layer-wise_aggregation,FedLAMA_AAAI_23} employs a layer-wise model aggregation method to reduce communication cost. \cite{AAAI25_FedAA} utilizes a compression technique based on model parameters of the last hidden layer for aggregation.
\cite{FedLF} designs a method to calculate a
layer-wise fair direction. \cite{IJCAI_Shapley} leverages Shapley value approximations, derived from the gradients of the last layer, to guide a weighted aggregation scheme. In contrast, we design a novel framework to alleviate bias by aligning client embedding vectors obtained from the last layer of the feature extractor. 

\subsection{Representation Fairness} 
Some recent studies have focused on representation fairness in FL. For example, \cite{Comp_CVPR_2025} envisages an augmentation technique to mitigate feature shift by injecting statistical information from the entire federation into each client's data. \cite{IEEE_trans_classifier} employs representation unification and prototypical mix-up to reduce bias in FL with long-tailed data. \cite{REp_Bias_NIPS_21} introduces a method to adjust the classifier using virtual representations. Similarly, \cite{Bias_CVPR_23} proposes bias-eliminating augmenters at each
client with the goal of generating bias-conflicting samples to eliminate local data biases. \cite{NEURIPS2023_2e0d3c6a} addresses representation degeneration by decomposing local representations into a global component and a client-specific bias term regulated via a mean regularization mechanism. 

Whilst these techniques have significantly advanced representation fairness in FL, they typically treat representation learning and aggregation as disjoint processes, i.e., without considering representation similarity in the aggregation logic. As a result, global models may still reflect skewed or incoherent representations, particularly under highly heterogeneous settings. In comparison, our method uniquely combines embedding alignment, based on contrastive loss and knowledge distillation, with historical-participation-aware weighting to jointly promote fair and coherent global representations across dynamically participating clients. 

\subsection{Collaborative Fairness}
Collaborative fairness in FL \cite{AAAI25_FedAA} ensures that participants who contribute more receive higher 
rewards in contrast to those with lower contributions. Several studies have addressed this aspect. For instance, \cite{Colab_Fair_KDD} maintains bounded collaborative fairness by assigning submodels based on client contributions and fair aggregation of low-frequency neurons. \cite{Colab_fair_CVPR} estimates client contributions in both gradient and data space using gradient direction differences and auxiliary model prediction errors to guide aggregation. Some other works \cite{IJCAI_Shapley, Colab_fair_NEURIPS2021} propose a reward mechanism based on the Shapley value to proportionally reward each client based on their respective contribution. In comparison, we propose a contribution-aware mechanism that assigns aggregation weights to clients based on their embedding alignment with a global model and inverse of participation frequency over a dynamic sliding window. This dual-component weighting establishes a reward-penalty mechanism to ensure that clients contributing with valuable and consistent information are prioritized during aggregation.

\begin{figure*}[t]
\centering
\includegraphics[width=0.95\textwidth]{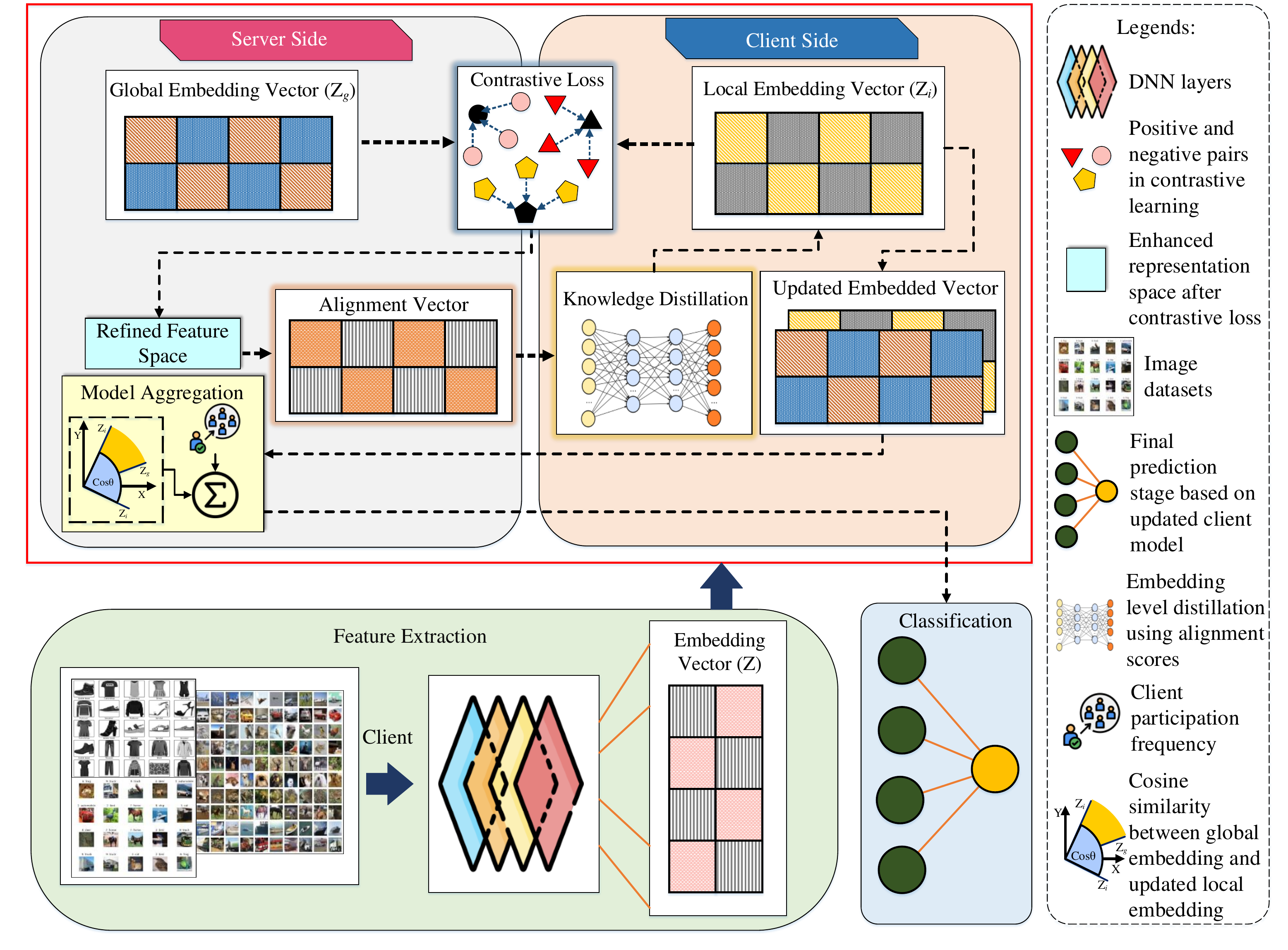} 
\caption{Architecture overview of CoRe-Fed framework. Clients perform local training and extract embedding vectors using DNN-based feature extractors. A server refines embedding space using contrastive learning over client and global embeddings. An alignment vector then guides knowledge distillation to adjust client representations towards a global semantic structure. Finally, contribution-aware aggregation is performed based on clients' embedding alignment and participation frequency.}
\label{Fig: Architecture Diagram}
\end{figure*}

\section{Methodology}

In this section, we describe the overall architecture and the technical details of our proposed framework, 
illustrated in Figure \ref{Fig: Architecture Diagram} and Algorithm \ref{alg:corefed}.
At beginning, each client initiates local training by performing feature extraction using DNNs on its private dataset to extract local embedding vector $Z_i$ (Line \ref{Step: Train data from dataset}). Subsequently, the server computes a global embedding vector $Z_g$ and evaluates contrastive loss between each client's embedding vector and a global embedding vector to quantify representational similarity across clients (Lines \ref{Step: Compute global Embedding}--\ref{Step: Normalize and compute contrastive loss}). Based on these relations, a server then constructs an alignment vector that guides knowledge distillation to align client embeddings towards a global embedding by explicitly reducing the distance between them (Lines \ref{Step: compute Alignment Score}--\ref{Step: Update Embedding using Alignment vector and knowledge distillation}). After knowledge distillation, the server executes a fairness-aware aggregation scheme based on two parameters; (i) a client's participation frequency score $\tau$ based on their sliding participation window and (ii) a representation similarity score $\alpha_i$ for each client that is derived from cosine similarity between refined embeddings and global embedding (Lines \ref{Step: Determine sliding window}--\ref{Step: Compute representation similarity}). Finally, these aggregation parameters are leveraged to compute contribution weights that balance clients' participation frequency with high representational quality (Line \ref{Step: Compute updated weight using participation frequency and similarity}). 
The remainder of this section is organized 
into two parts,
problem formulation and the technical components of the CoRe-Fed framework.

{
\subsection{Problem Formulation and Local Objective}
We consider FL setting involving a central server that orchestrates training across a set of $m$ clients, $\mathcal{C} = \{1, \dots, m\}$, where each client $i$ possesses a private, locally stored dataset $D_i = \{(x_{ij}, y_{ij})\}_{j=1}^{n_i}$. The goal is to collaboratively optimize a global model without sharing raw data. The global model $\omega \in \mathbb{R}^d$ comprises a feature extractor $\phi_\omega$ and a predictor module. Each input is processed as:
\begin{equation}
f_\omega(x) = \text{Predictor}(\phi_\omega(x)),
\label{Eq: Predictor Forward Pass}
\end{equation}
where, $\phi_\omega(x)$ is the embedding vector from the feature extractor. The local training loss on client $i$ is defined as:
\begin{equation}
F_i(\omega) = \frac{1}{n_i} \sum_{j=1}^{n_i} \ell(f_\omega(x_{ij}), y_{ij})
\label{Eq: FL Training Loss}
\end{equation}

\noindent The global learning objective is the weighted average of local losses and is calculated as:
\begin{equation}
\min_{\omega} \sum_{i=1}^m \frac{n_i}{n} F_i(\omega)
\label{Eq: Global Objective}
\end{equation}
where, $n = \sum_i n_i$ is the total data count across all clients.

\subsection{The CoRe-Fed Framework}
Building on the above formulation, CoRe-Fed comprises two synergistic modules: (i) representation alignment and (ii) fairness-aware aggregation, each detailed below.

\subsubsection{Embedding Extraction and Normalization.}

To represent client-specific data in a feature space, each client extracts embeddings from its local samples using a DNN encoder. The embedding vector for a client $i$ is computed by averaging the L2-normalized feature embeddings across its local dataset:
\begin{equation}
z_i = \frac{1}{n_i} \sum_{j=1}^{n_i} \frac{\phi_\omega(x_{ij})}{\|\phi_\omega(x_{ij})\|}
\label{Eq: Client-wise Normalized Embedding}
\end{equation}
A global feature embedding is then obtained at the server by averaging over a subset of participating clients $\mathcal{C}_t$:
\begin{equation}
z_g = \frac{1}{|\mathcal{C}_t|} \sum_{i \in \mathcal{C}_t} z_i
\label{Eq: Global Reference Embedding Vector}
\end{equation}

\subsubsection{Contrastive Learning for Representation Consistency.}
Recently, contrastive learning has gained prominence in FL as a powerful approach for improving representation quality under data heterogeneity. For instance, 
\cite{AAAI_2025_ContrastiveLearning_graph} leverages contrastive objectives to capture meaningful graph structures.   
\cite{AAAI_2025_ContrastiveLearning} applies contrastive loss to learn shared and personalized dynamics.  
\cite{AAAI_2025_Contrastive} aligns multiple embeddings to address cross‑modal inconsistencies.  
\cite{NIPS2022fairvfl} combines contrastive loss with adversarial training to reduce representational unfairness. Inspired by these works, we apply contrastive representation alignment to enforce semantic consistency between local and global embeddings in a federated setting. 
We adopt the temperature-scaled InfoNCE (NT-Xent) loss to align each client embedding $z_i$ with the global embedding $z_g$ while contrasting it against other client embeddings $\{z_l\}_{l \ne i}$:
\begin{equation}
\mathcal{L}_{\text{contrast}}^{(i)} = -\log \frac{\exp(\text{cos}(z_i, z_g)/\tau_c)}{\sum\limits_{l \ne i} \exp(\text{cos}(z_i, z_l)/\tau_c)}
\label{Eq: Contrastive loss}
\end{equation}
Here, $\tau_c$ is a temperature parameter and $\text{cos}( )$ denotes cosine similarity, ascertained as:
\begin{equation}
\text{cos}(z_i, z_l) = \frac{z_i^\top z_l}{\|z_i\| \cdot \|z_l\|}
\label{Eq: Cosine Sim b/w Vectors}
\end{equation}

\subsubsection{Embedding-Level Knowledge Distillation.}

To reduce representational divergence, client embeddings are softly aligned towards a global embedding structure via embedding-level knowledge distillation, 
wherein, each client updates its representation using an alignment vector. This vector serves as a reference for embedding-level knowledge distillation, thus encouraging representation alignment while preserving local features. The updated embedding vector for client \( i \) is:
\begin{equation}
\tilde{z}_i = z_i + \beta \cdot (\tilde{z}_g^{(i)} - z_i)
\label{Eq: Knowledge Distillation}
\end{equation}
where, \( \beta \in [0,1] \) is a knowledge distillation coefficient, and \( \tilde{z}_g^{(i)} \in \mathbb{R}^d \) is the alignment vector for client \( i \) computed as:
\begin{equation}
\tilde{z}_g^{(i)} = A_i \cdot z_g
\label{Eq: Alignment Vector}
\end{equation}
here, $ A_i = \cos(z_i, z_g)$  is the alignment score, computed based on cosine similarity with a server-side global embedding that captures the shared semantic structure. The resulting alignment vector \( \tilde{z}_g^{(i)} \) provides a refinement signal that guides each client's embedding update in the direction of the global semantic structure. This distillation encourages each client to preserve its unique structure while adapting to the global semantic space.

\subsubsection{Participation Frequency Estimation.}

To ensure absent clients' fairness, CoRe-Fed estimates each client's participation frequency over a recent sliding window following the approach proposed in \cite{FedLF,wang2021FedFV}. Specifically, for client \( i \), the frequency \( f_i \) is computed as the number of rounds where \( i \in C_r \) for \( r = t - \tau + 1 \) to \( t \) and \( C_r \) is the set of online clients in round \( r \). The window length \( \tau \) is dynamically defined as \( \tau = \frac{M}{|C_t|} \), where \( M \) is the number of distinct clients that have participated so far and \( |C_t| \) is the number of online clients in the current round. This formulation captures both short-term participation and inactivity.
Each client $i$'s recent participation frequency $f_i$ over a dynamic window $\tau$ is computed as:
\begin{equation}
    f_i = \frac{1}{\tau} \sum_{r = t - \tau + 1}^{t} \mathbb{I}[i \in \mathcal{C}_r]\
    \label{Eq: Participation Frequency}
\end{equation}
where, $\mathbb{I}[\cdot]$ is the indicator function that evaluates to 1 if the client participated in round $r$.
\setlength{\algomargin}{1.6em} 
\begin{algorithm}[!tb]
\DontPrintSemicolon
\caption{CoRe-Fed: Collaborative Representation Fairness in Federated Learning}
\label{alg:corefed}
\KwIn{Initial global model $\omega^0$, total rounds $T$, local epochs $E$, client set $\mathcal{C} = \{1, \dots, N\}$, temperature $\tau_c$, distillation weight $\beta$, scaling factor $k$, fairness exponent $\gamma$}
\KwOut{$\omega^T$}
\For{$t = 1$ to $T-1$}{
    \textbf{// Server-side:} Select online clients $\mathcal{C}_t \subseteq \mathcal{C}$\; \label{Step: Select Online Clients}
    Broadcast global model $\omega^t$ to all $i \in \mathcal{C}_t$\; \label{Step: Broadcast Global Model}

    \textbf{// Client-side (in parallel):}\;
    \For{$i \in \mathcal{C}_t$}{
        Initialize local model $\omega_i^t \gets \omega^t$\; 
        \For{$e = 1$ to $E$}{
            Train $\omega_i^t$ on local data $\mathcal{D}_i$\; \label{Step: Train data from dataset}
        }
        Extract local embedding $z_i$ by Eq. (\ref{Eq: Client-wise Normalized Embedding})\; \label{Step: Extraxt Local Embeddings}
        Send $(\omega_i^t, z_i)$ to server\; \label{Step: Send Embeddings and weights to Server}
    }

    \textbf{// Server-side: Contrastive Alignment}\;
    Compute global embedding $z_g$ by Eq. (\ref{Eq: Global Reference Embedding Vector})\; \label{Step: Compute global Embedding} 
    Compute contrastive loss $\mathcal{L}_{\text{contrast}}$ between normalized $z_i$ and $z_g$ by Eq. (\ref{Eq: Contrastive loss})\; \label{Step: Normalize and compute contrastive loss}

    \ForEach{$i \in \mathcal{C}_t$}{
        Compute alignment vector $\tilde{z}_g^{(i)} = A_i \cdot z_g$\; \label{Step: compute Alignment Score}
        Update embedding using Knowledge Distillation Eq. (\ref{Eq: Knowledge Distillation});
        \label{Step: Update Embedding using Alignment vector and knowledge distillation}
    }

    \textbf{Server-side: Fairness-aware Aggregation}\;
    Determine sliding window by $\tau = \frac{M}{|C_t|}$ \; \label{Step: Determine sliding window}
    \ForEach{$i \in \mathcal{C}_t$}{
        Compute $f_i = \frac{1}{\tau} \sum_{r = t - \tau + 1}^{t} \mathbb{I}[i \in \mathcal{C}_r]$\; \label{Step: Compute participation frequency}
        Compute $\rho_i = \cos(\tilde{z}_i, z_g)$\; \label{Step: Compute representation similarity}
        Compute $w_i = \frac{ \left( \frac{1}{f_i} \right)^\gamma \cdot \sigma(k \cdot \rho_i)}{ \sum_{l \in \mathcal{C}_t} \left( \frac{1}{f_l} \right)^\gamma \cdot \sigma(k \cdot \rho_l) }$\; \label{Step: Compute updated weight using participation frequency and similarity}
    }
    Normalize weights: $w_i \gets \frac{w_i}{\sum_l w_l}$\; \label{Step: Normalize final weights}
    Aggregate: $\omega^{t+1}$ by Eq. (\ref{Eq: Final Model Aggregation})\;
}
\KwRet{$\omega^T$}
\end{algorithm}

\subsubsection{Sigmoid‑Modulated Fairness Weighting.}

To foster equitable contribution during optimization, CoRe-Fed combines participation frequency and representational alignment into a unified weighting scheme. This approach rewards under‑represented and semantically aligned clients while softly penalizing over‑represented or poorly aligned ones. If \(f_i\) is participation frequency of client \(i\) over a recent sliding window (Eq.~\ref{Eq: Participation Frequency}) and \(\rho_i = \text{cos}(\tilde{z}_i, z_g)\) is the cosine similarity between its refined embedding \(\tilde{z}_i\) and the global embedding \(z_g\), then the aggregation weight is computed as:
\begin{equation}
     w_i = \frac{ \left( \tfrac{1}{f_i} \right)^\gamma \cdot \sigma(k \cdot \rho_i)}{ \sum_{l \in \mathcal{C}_t} \left( \tfrac{1}{f_l} \right)^\gamma \cdot \sigma(k \cdot \rho_l) }
  \label{Eq: Sigmoid Modulated Fairness}
\end{equation}
where, $\gamma$ is the fairness exponent and $k$ controls slope of sigmoid sensitivity to alignment score. The sigmoid function $\sigma$ can be defined as ${\sigma(x)=1/(1+e^{-x})}$.
As depicted in Eq.~(\ref{Eq: Sigmoid Modulated Fairness}), the participation factor ${\left(\tfrac{1}{f_i}\right)^\gamma}$ amplifies the influence of under-represented clients while softly down-weighting frequent ones. In contrast, the alignment factor ${\sigma(k \cdot \rho_i)}$ favors clients that better aligns with the global model and penalizes misaligned ones.

\subsubsection{Fairness-Aware Aggregation with Gradient Reuse.}
To incorporate influence of temporarily inactive clients' fair contribution, we follow \cite{FedLF} to reuse historical gradients within a sliding window of \(\tau\) rounds. For each client \(i\), the reused gradient \(\hat{g}_i\) is defined as:
\begin{equation}
  \hat{g}_i =
\begin{cases}
g_i^{(t)} & \text{if } i \in \mathcal{C}_t \\
g_i^{(t_i)} & \text{if } t - t_i \leq \tau \\
0 & \text{otherwise}
\end{cases}
\label{Eq: Gradient Use History}
\end{equation}
where, \(g_i^{(t)}\) is the current gradient, and \(g_i^{(t_i)}\) is the last known gradient from client \(i\) within the \(\tau\)-round window. These gradients are combined with fairness-aware weights \(w_i\) to update the global model:
\begin{equation}
 \omega_{t+1} = \omega_t - \eta \cdot \sum_{i \in \mathcal{C}_t} w_i \cdot \hat{g}_i 
 \label{Eq: Final Model Aggregation}
\end{equation}

\section{Experiments and Discussions}

\subsection{Evaluation Metrics}
To provide a comprehensive evaluation of fairness, we adopt two widely used metrics, i.e., cosine similarity-based angular distance \cite{FedMDFG_AAAI_2023,wang2021FedFV} and Manhattan distance \cite{Huang_2023_ICCV}. In particular, we compute the angular cosine similarity distance between a client model $\boldsymbol{\phi}_i$ and a global model $\boldsymbol{\phi}^*$ as $D_{Cosine} = \arccos\left( \frac{ \boldsymbol{\phi}_i \cdot \boldsymbol{\phi}^* }{ \| \boldsymbol{\phi}_i \| \, \| \boldsymbol{\phi}^* \| } \right)$ to evaluate representation fairness by measuring their directional alignment. In contrast, Manhattan distance $D_{Manhattan} = \sum_{j=1}^{d} \left| \phi_i^{(j)} - \phi^{*(j)} \right|$ quantifies the model performance fairness by capturing the magnitude of parameter divergence between a client model and a global model, thereby reflecting how much each client benefits or deviates from a shared global model.

\begin{table*}[ht] 
\centering
\scalebox{0.9}{
\begin{tabular}{l|ccc|ccc}
\toprule
\multirow{2}{*}{\textbf{Algorithm}} 
& \multicolumn{3}{c|}{\textbf{FMNIST}} 
& \multicolumn{3}{c}{\textbf{CIFAR-10}} \\
\cmidrule(lr){2-4} \cmidrule(lr){5-7}
& Accuracy $\uparrow$ & $D_{Cosine} \downarrow$ &  $D_{Manhattan} \downarrow$ 
& Accuracy $\uparrow$ & $ D_{Cosine} \downarrow$ & $ D_{Manhattan} \downarrow$ \\
\midrule
FedRDN$^\dag$ \cite{Comp_CVPR_2025}    & 0.870 & 0.746 & 116.8 & 0.569 & 1.077 & 180.9 \\
FedMDFG \cite{FedMDFG_AAAI_2023}       & 0.874 & 0.587 & 88.1 & 0.681 & 0.766 & 116.3 \\
FedMGDA+ \cite{MDGA+comp}              & 0.849 & 0.421 & 79.5 & 0.549 & 0.719 & 48.4 \\
Ditto \cite{li2021dittofairrobustfederated} & 0.862 & 0.536 & 106.5 & 0.663 & 1.251 & 104.2 \\
qFedAvg \cite{li2020fair_ICLR}         & 0.884 & 0.401 & 76.2 & 0.628 & 0.702 & 52.9 \\
\midrule
CoRe-Fed (Ours) & \textbf{0.891} & \textbf{0.294} & \textbf{73.5} & \textbf{0.722} & \textbf{0.430} & \textbf{36.0} \\
\bottomrule
\end{tabular}
}
\caption{Comparison of mean test accuracy, angular cosine similarity distance $(D_{Cosine})$, and Manhattan distance $(D_{Manhattan})$ on FMNIST and CIFAR-10 with Dir (0.5), batch size 50, 1000 rounds, and 20 out of 100 clients per round.}
\label{tab:algorithm_comparison}
\end{table*}

\begin{figure*}[ht]
    \centering
    \includegraphics[width=0.85\linewidth]{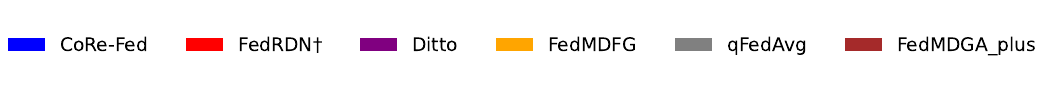}
    \begin{subfigure}[t]{0.24\linewidth}
        \centering
        \includegraphics[width=\linewidth]{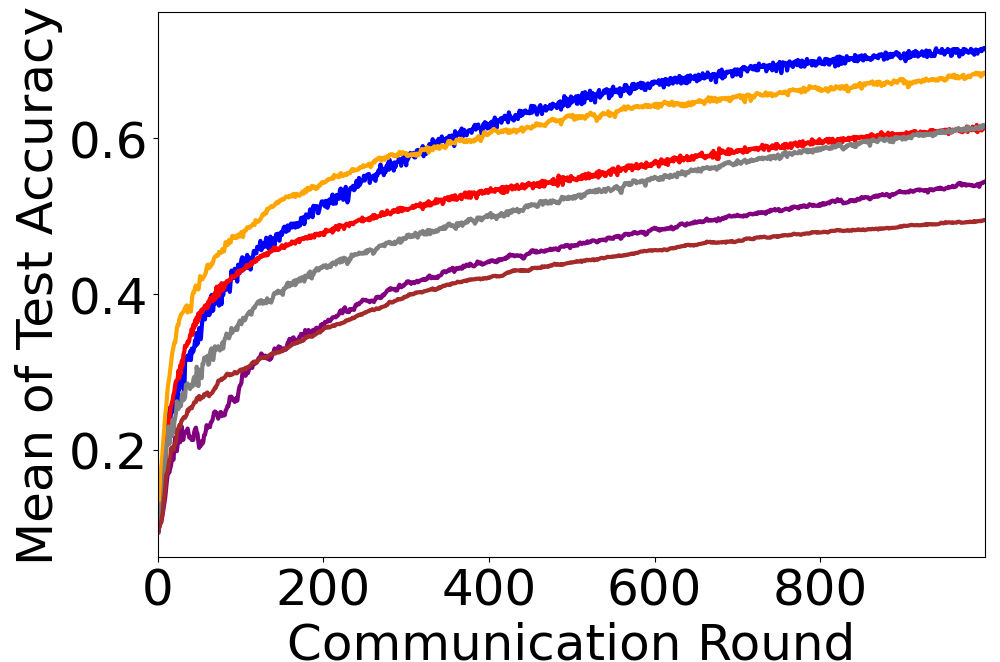}
        \label{Fig: CNN Acc batch 200}
    \end{subfigure}
    \begin{subfigure}[t]{0.24\linewidth}
        \centering
        \includegraphics[width=\linewidth]{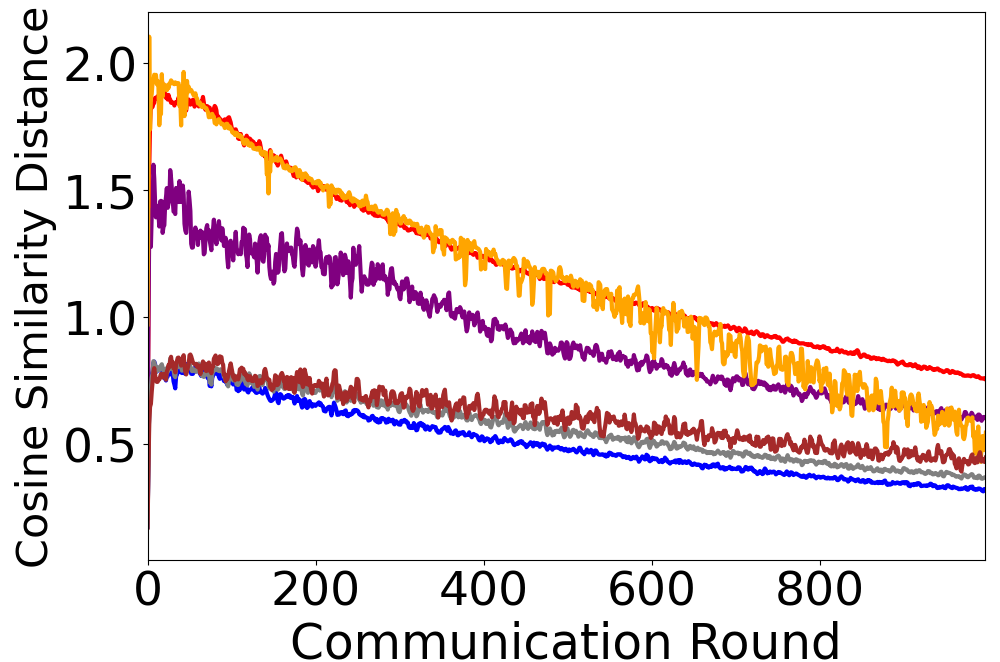}
        \label{Fig: CNN Cosine Similarity batch 200}
    \end{subfigure}
    \begin{subfigure}[t]{0.24\linewidth}
        \centering
        \includegraphics[width=\linewidth]{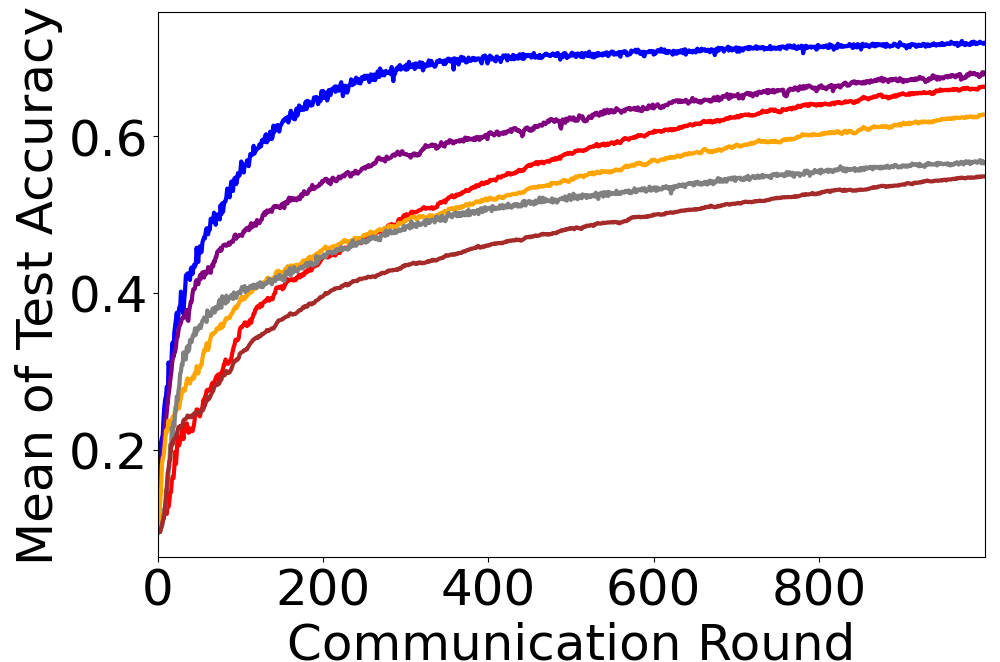}
        \label{Fig: CNN Acc batch 50}
    \end{subfigure}
    \begin{subfigure}[t]{0.24\linewidth}
        \centering
        \includegraphics[width=\linewidth]{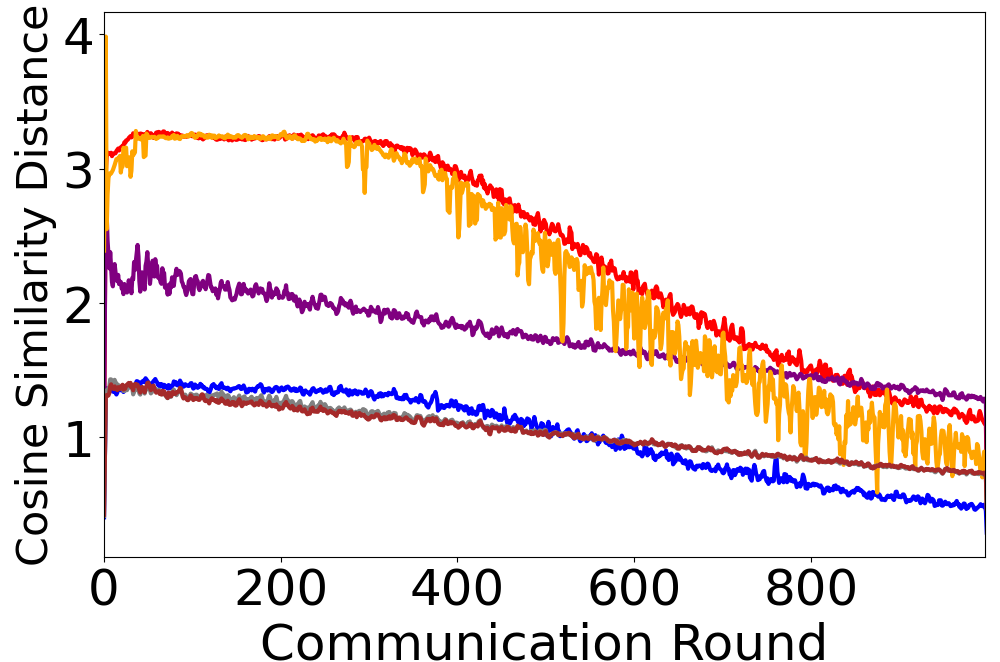}
        \label{Fig: CNN Cosine Similarity Batch 50}
    \end{subfigure}

    {\small
    \makebox[0.48\linewidth][c]{(a) Batch size 200}  \label{Fig: Overall model Batch 200}
    \hfill
    \makebox[0.48\linewidth][c]{(b) Batch size 50}
    } \label{Fig: Overall model Batch 50}

    \caption{Mean test accuracy and fairness on CIFAR-10 with batch sizes 50 and 200; 20 out of 100 clients online per round.}
    \label{Fig: Training rounds}
\end{figure*}

\subsubsection{Baselines and Hyper-parameters.}
Our evaluation benchmarks against several state-of-the-art fairness-based FL algorithms including q‑FedAvg, \cite{li2020fair_ICLR}, FedMGDA+ \cite{MDGA+comp}, Ditto \cite{li2021dittofairrobustfederated}, FedMDFG \cite{FedMDFG_AAAI_2023}, and FedRDN \cite{Comp_CVPR_2025}. We modify 
FedRDN$^\dag$ by incorporating a small constant $\epsilon = 10^{-6}$ into the standard deviation and mean calculations to prevent division by zero and ensure model stability. 
We follow the experimental settings of \cite{FedMDFG_AAAI_2023}, wherein all clients perform local training using Stochastic Gradient Descent with local epoch $E = 1$. The learning rate is set to $\eta = 0.1$ with a decay factor of 0.999 per round. All results are averaged over three runs with different random seeds. The hyper-parameters used for each technique are summarized in Table~\ref{tab:hyperparams}. 
\begin{table}[!ht]
\centering
\scalebox{0.95}{
\begin{tabular}{l|l}
\hline
\textbf{Method} \rule{0pt}{2.2ex} & \textbf{Hyper-parameters} \rule{0pt}{2.2ex} \\ \hline
qFedAvg    & $q \in \{0.1, 1.0\}$ \\
Ditto      & $\lambda \in \{0.1, 1.0\}$ \\
FedMGDA+   & $e \in \{0.1, 1.0\}$ \\
FedMDFG    & $\theta \in \{\frac{\pi}{180}\} ,\ s \in \{5, 3, 1\}$ \\ 
CoRe-Fed  &  $\gamma \in \{0.5,2\}$,\; $k \in \{0.5, 2\}$,\\ & $\beta$ = 0.5, \;$\tau_c$ = 0.07 \\ \hline
\end{tabular}}
\caption{Hyper-parameters of different methods}
\label{tab:hyperparams}
\end{table}
\subsubsection{Datasets and Models.}
We evaluate the performance of mentioned baselines on two image classification datasets, i.e., Fashion MNIST (FMNIST) \cite{fashionMNIST} and CIFAR-10 \cite{CIFAR10/100}.
To simulate heterogeneous client distributions, we adopt the Dirichlet-based non-IID partitioning strategy from \cite{Dir_settings}, where data is distributed among $m$ clients according to a Dirichlet distribution with concentration parameter $\alpha$. When $\alpha < 1$, the data distribution becomes increasingly skewed, leading to substantial label imbalance and non-uniform data volumes across clients. We adopt model architectures and configurations consistent with \cite{wang2021FedFV, FedMDFG_AAAI_2023}, i.e., a Multilayer Perceptron (MLP) for FMNIST and Convolutional Neural Network (CNN) with two convolutional layers for CIFAR-10.

\subsection{Performance and Fairness}
The experimental results of CoRe-Fed vis-\`a-vis several state-of-the-art baselines across FMNIST and CIFAR-10 datasets are presented in Table~\ref{tab:algorithm_comparison}. It can be observed that CoRe-Fed consistently achieves the highest accuracy with relative gains of 0.8\% and 6.0\%, respectively, over the best-performing baselines, i.e., qFedAvg for FMNIST and FedMDFG for CIFAR-10. The observed improvements are driven by CoRe-Fed’s integration of embedding-level alignment and contribution-aware aggregation, which enhance local representation quality and ensure equitable contributions during aggregation. Moreover, in terms of representation fairness, CoRe-Fed reports the lowest angular misalignment with respective reductions of 26.7\% and 43.9\% compared to the best methods. This demonstrates its ability to align client representations more closely with the global model by integration of an alignment-aware weighting, which favors clients whose updates are directionally consistent with the global objective. 
Similarly, for model performance fairness, it records the lowest divergence with respective improvements of 3.5\% and 69\% compared to qFedAvg and FedMDFG. This suggests that CoRe-Fed not only reduces representation bias but also minimizes performance bias by ensuring that updates from all clients contribute proportionately and meaningfully. 

As evident from Figure \ref{Fig: Training rounds}, CoRe-Fed consistently achieves superior test accuracy, enhanced fairness, faster convergence, and improved generalization under batch sizes of 50 and 200, respectively.  The angular cosine similarity metric further confirms reduced representational divergence throughout training. Particularly, with the smaller batch size where client updates are inherently noisier, some baseline methods, e.g., FedMGDA+ and FedRDN, exhibit instability. In contrast, CoRe-Fed maintains stable and smooth learning dynamics,
demonstrating robustness to both sampling noise and client heterogeneity. Moreover, we evaluate per client test accuracy on the FMNIST dataset with Dirichlet $\alpha \in \{0.1, 0.5\}$. As illustrated in Figure \ref{Fig: PerClient Acc}, CoRe-Fed consistently outperforms all baseline methods achieving both high average accuracy and low inter-client variance. 
\begin{figure}[!ht]
    \centering
    \begin{subfigure}[t]{0.48\linewidth}
        \centering
        \includegraphics[width=\linewidth]{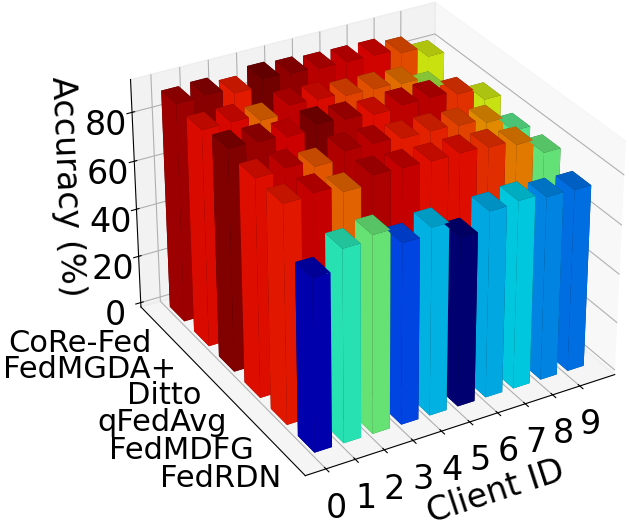}
        \caption{}
        \label{Fig: PerClient Acc FMNIST 0.5}
    \end{subfigure}
    \begin{subfigure}[t]{0.48\linewidth}
        \centering
        \includegraphics[width=\linewidth]{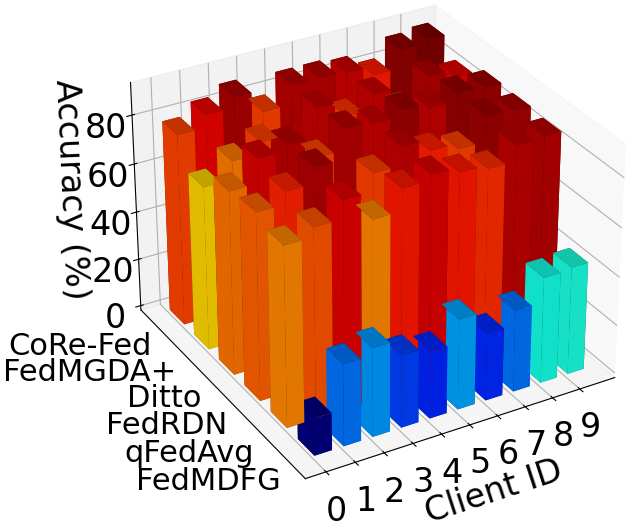}
        \caption{}
        \label{Fig: PerClient Acc FMNIST 0.5}
    \end{subfigure}
    \caption{The average per client test accuracy on FMNIST with batch size 200; 100\% of 10 clients online per round.}
    \label{Fig: PerClient Acc}
\end{figure}

\subsection{Ablation Experiments}
To assess the effectiveness of different key components in our proposed framework, we conduct ablation experiments across three scenarios, i.e.,
(a) \textbf{Co-Fed:} In this scenario, we exclude both contrastive learning and knowledge distillation components thus retaining only the contribution-aware aggregation strategy; (b) \textbf{Re-Fed:} In this scenario, we eliminate contribution-aware aggregation and instead use standard aggregation technique, i.e., FedAvg, while retaining contrastive learning and knowledge distillation to address representation bias; (c) \textbf{CoRe-Fed:} This scenario combines both contribution-aware aggregation and representation alignment.
As evident from Figure~\ref{Fig: Ablation Experiments}, Re-Fed reduces representation bias through enhanced local training but lacks contribution-aware aggregation, thereby allowing dominant clients to overshadow others during global aggregation. This results in reduced accuracy and elevated fairness distances thus demonstrating that representation fairness alone is insufficient. Conversely, Co-Fed incorporates contribution-aware aggregation but omits representation-level adjustments, which leads to marginal fairness gains at the cost of model performance. In contrast, CoRe-Fed consistently outperforms both ablated variants achieving higher accuracy and lower cosine and Manhattan distances thus indicating more balanced and fair client contributions.
\begin{figure}[!ht]
    \centering
    \begin{subfigure}[t]{0.325\linewidth}
        \centering
        \includegraphics[width=\linewidth]{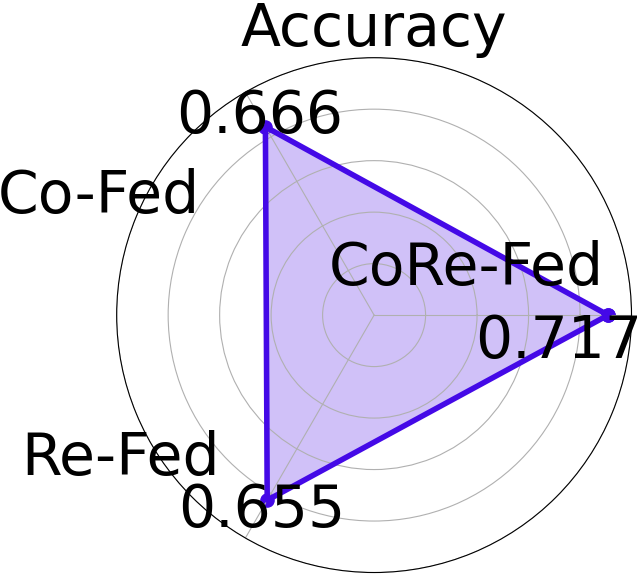}
        \caption{}
        \label{Fig: Ablation Accuracy}
    \end{subfigure}
    \begin{subfigure}[t]{0.325\linewidth}
        \centering
        \includegraphics[width=\linewidth]{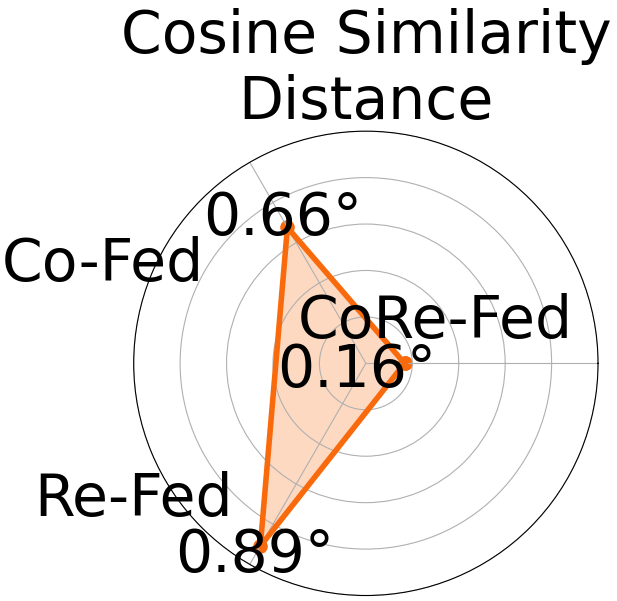}
        \caption{}
        \label{Fig: Ablation Cosine Similarity}
    \end{subfigure}
     \begin{subfigure}[t]{0.325\linewidth}
        \centering
        \includegraphics[width=\linewidth]{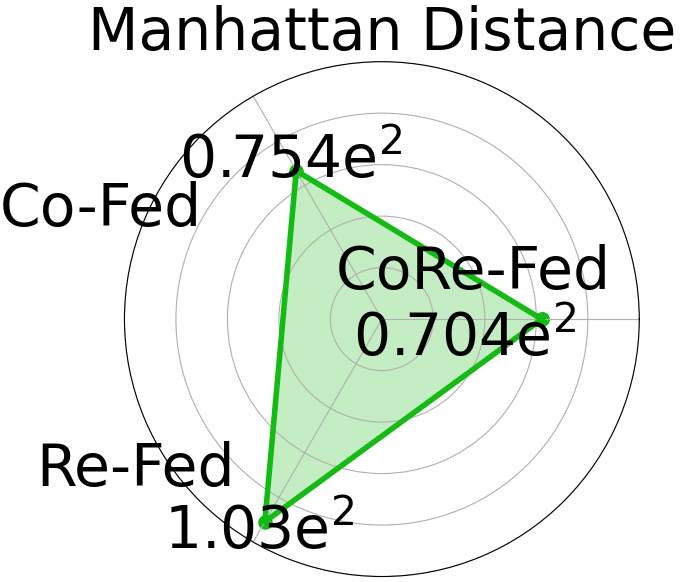}
        \caption{}
        \label{Fig: Ablation Manhattan}
    \end{subfigure}
    \caption{Ablation experiments comparing three scenarios.}
    \label{Fig: Ablation Experiments}
\end{figure}
 \subsubsection{The Trade-off between Hyper-parameters $\gamma$ and $k$. } The ablation results in Table \ref{tab: ablation_k_tau} demonstrate the trade-offs driven by the fairness-aware aggregation hyper-parameters $k$ and $\gamma$. As observed, increasing $k$ from 0.5 to 2.0 sharpens the sigmoid weighting function and strengthens alignment between clients and the global model. Meanwhile, lowering $\gamma$ from 2.0 to 0.5 reduces the emphasis on infrequent clients thereby easing penalization on frequent ones. The latter adjustment yields consistent improvements across all metrics, i.e., accuracy increases by 0.81\%, cosine-based angular distance decreases by 0.16\%, and Manhattan distance reduces by 0.32\%. While a high $\gamma$ promotes fairness by amplifying under-represented clients, it may introduce outdated updates and hinder convergence. In particular, overly amplifying the influence of non-participating or rarely participating clients may introduce divergent updates into the global model thus degrading stability and representational coherence. 
\begin{table}[!ht]
\centering
\scalebox{0.82}{
\begin{tabular}{lccc}
\toprule
\textbf{Scenario} & \textbf{Acc $\uparrow$} & \textbf{$D_{Cosine} \downarrow$} & \textbf{$D_{Manhattan}\downarrow$} \\
\midrule
$k=0.5,\ \gamma=2.0$     & 0.861  & 0.12704 & 52.88 \\
$k=2.0,\ \gamma=0.5$     & 0.868  & 0.12684 & 52.71 \\
\midrule
\textbf{$\Delta (\%)$}        & +0.81\% & +0.16\% & +0.32\% \\
\bottomrule
\end{tabular} 
}
\caption{The trade-off between fairness hyper-parameters.}
\label{tab: ablation_k_tau}
\end{table}
\section{Conclusion}
This paper introduces CoRe-Fed, a unified federated optimization framework designed to simultaneously maintain two distinct but interconnected notions of fairness, 
i.e., collaborative and representation fairness. By integrating embedding-level alignment and contribution-aware aggregation, CoRe-Fed fosters semantic coherence and equitable participation, thereby enhancing both generalization and fairness across clients. Our findings open several avenues for future exploration.  One promising direction is the design of personalization-aware fairness objectives that effectively balance global collaboration with individual client preferences. 

\section*{Acknowledgments}
The authors sincerely acknowledge the generous support of Macquarie University for funding the research via its `International Research Excellence Award (Allocation No. 20235578)'.

\end{document}